# Adaptive Importance Sampling for Estimation in Structured Domains

**Luis E. Ortiz**
Computer Science Department
Brown University
Box 1910
Providence, RI 02912 USA
leo@cs.brown.edu

**Leslie Pack Kaelbling**
Artificial Intelligence Laboratory
Massachusetts Institute of Technology
545 Technology Square
Cambridge, MA 02139 USA
lpk@ai.mit.edu

## Abstract

Sampling is an important tool for estimating large, complex sums and integrals over high-dimensional spaces. For instance, importance sampling has been used as an alternative to exact methods for inference in belief networks. Ideally, we want to have a sampling distribution that provides optimal-variance estimators. In this paper, we present methods that improve the sampling distribution by systematically adapting it as we obtain information from the samples. We present a stochastic-gradient-descent method for sequentially updating the sampling distribution based on the direct minimization of the variance. We also present other stochastic-gradient-descent methods based on the minimization of typical notions of *distance* between the current sampling distribution and approximations of the target, optimal distribution. We finally validate and compare the different methods empirically by applying them to the problem of action evaluation in influence diagrams.

## 1 INTRODUCTION

Often, we are interested in computing quantities involving large sums, such as expectations in uncertain, structured domains. For instance, belief inference in *Bayesian networks (BNs)* requires that we sum or marginalize over the remaining variables that are not of interest. Similarly, in order to solve the problem of action selection in *influence diagrams*, we sum over the variables that are not observed at the time of the decision in order to compute the value of different action choices.

We can represent the uncertainty in structured environments using a BN. A BN allows us to compactly define a joint probability distribution over the relevant variables in a domain. It provides a graphical representation of the distribution by means of a directed acyclic graph (DAG). It defines locally a conditional probability distribution for each relevant variable, represented as a node in the graph, given the state of its parents in the graph. This decomposition can help in the evaluation of the sums. However, due to factors regarding the connectivity of the graph, in general this is not sufficient to allow an efficient computation of the exact value of the sums of interest.

Sampling provides an alternative tool for approximately computing these sums. Sampling methods have been proposed as an alternative to exact methods for such problems. In particular, *importance sampling* (see Geweke [1989], and the references therein) has been applied to the problem of belief inference in BNs [Fung and Chang, 1989, Shachter and Peot, 1989] and action selection in IDs (see Charnes and Shenoy [1999] and the references therein, and Ortiz and Kaelbling [2000]). In its simpler form, the importance-sampling distribution used is the "prior" distribution of the BN resulting from setting the value of the evidence. It has been noted early on that this sampling distribution is far from optimal in the sense that it provides estimates with larger variance than necessary [Shachter and Peot, 1989]. For instance, the optimal sampling distribution in the case of belief inference is to sample the unobserved variables from the posterior distribution over them given the observed evidence. If we knew this distribution we would know the answer to the belief inference problem.

Several modifications have been proposed to improve the estimation of the simple importance sampling distribution discussed above, based on information obtained from the samples [Fung and Chang, 1989, Shachter and Peot, 1989, Shwe and Cooper, 1991]. In this paper, we propose methods to systematically and sequentially update the importance-sampling distribution. We view the updating process as one of learning a separate BN just for sampling. The learning objective is to minimize some error criterion. A stochastic-gradient method results from the direct minimization of the variance of the estimator with respect to the importance sampling distribution as an error function. Other stochastic-gradient methods result from minimizing error functions based on typical measures of the notion of *distance* between the current sampling distribution and approximations of the optimal sampling distribution.

## 2 DEFINITIONS

We begin by introducing some notation used throughout the paper. We denote one-dimensional random variables by capital letters and denote multi-dimensional random variables by bold capital letters. For instance, we denote a multi-dimensional random variable by $X$ and denote all its components by $(X_1, \ldots, X_n)$ where $X_i$ is the $i^{th}$ one-dimensional random variable. We use small letters to denote assignments to random variables. For instance, $X = x$ means that for each component $X_i$ of $X$, $X_i = x_i$. We denote the set of possible values that $X_i$ can take by $\Omega_{X_i}$ and the set of possible values that $X$ can take by $\Omega_X = \times_{i=1}^{n} \Omega_{X_i}$. We also denote by capital letters the nodes in a graph. We denote by $\text{Pa}(Y)$ the parents of node $Y$ in a directed graph.

We now introduce notation that will become useful during the description of the methods presented in this paper. We denote by the operator $\sum_Z$ the sum over the possible values of the individual variables forming $Z$, $\sum_{Z_1} \sum_{Z_2} \cdots \sum_{Z_{n_1}}$. For any function $h$ with variables $Z$ and $O$, the expression $h(Z, O)|_{O=o}$ stands for a function $f'$ over variables $Z$ that results from setting the values of $O$ in $h$ with assignment $o$ while letting the values for $Z$ remain unassigned. In other words, $f'(Z) = h(Z, O)|_{O=o} = h(Z, O = o)$. The notation $X = (Z, O)$ means that the variable $X$ is formed by all the variables that form $Z$ and $O$. That is, $X = (X_1, \ldots, X_n) = (Z_1, \ldots, Z_{n_1}, O_1, \ldots, O_{n_2}) = (Z, O)$, where $n = n_1 + n_2$. Note that we are assuming that the set of variables forming $Z$ and those forming $O$ are disjoint. The notation $Z \sim f$ means that the random variable $Z$ is distributed according to probability distribution $f$.

A *Bayesian network (BN)* is a graphical probabilistic model used to represent uncertainty in structured domains. It compactly represents the joint probability distribution over the relevant variables of the system of interest. It uses a *directed acyclic graph (DAG)* to represent the relationship between the relevant variables. A node in the graph represents a variable. The model defines a local conditional distribution $P(X_i \mid \text{Pa}(X_i))$ for each node or variable $X_i$ given its parents $\text{Pa}(X_i)$ in the graph. The joint distribution is then

$$P(X) = \prod_{i=1}^{n} P(X_i \mid \text{Pa}(X_i)).$$

For instance, we can define a BN on the graph given in Figure 1(a).

The inference problem in BNs is that of computing the posterior probability of an assignment to a subset of variables given evidence about another subset of variables in the system. Assume that the variables are discrete and their *sample spaces* or the possible values each variable can take are finite. In general, let $X = (Z, O)$ where $O$ is the set of variables of interest, $o$ is an assignment to it and $Z$ are the remaining variables. For this problem we want to compute probabilities of the kind

$$P(O = o) = \sum_Z P(Z, O = o).$$

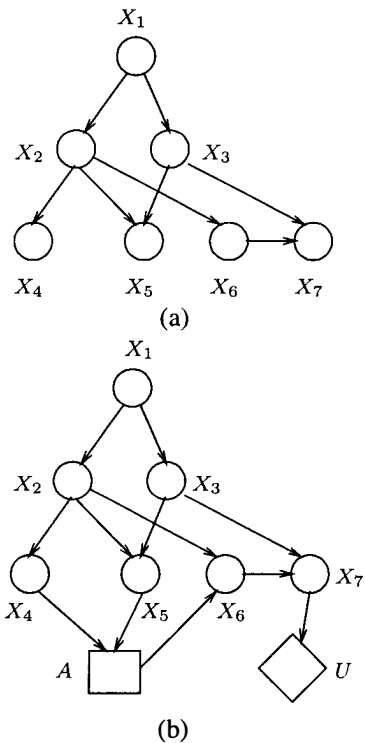

Figure 1: Example of (a) Bayesian network and (b) influence diagram.

Often, the local decomposition of the joint distribution still leads to the evaluation of sums over a large number of variables. In general, this problem is intractable [Cooper, 1990].

An *influence diagram (ID)* is a probabilistic model for decision-making under uncertainty. We can think of an ID as a BN with decision and utility nodes added. For instance, we can use our example BN to build an ID as shown in Figure 1(b). The square is a decision node. The diamond is a utility node. We now have potentially different joint distributions over the variables, for each action choice available. Assume for simplicity that there is a single decision node in the graph. The joint distribution over the variables, given the action choice $a$ assigned to the decision variable, is

$$P(X \mid A = a) = \prod_{i=1}^{n} P(X_i \mid \text{Pa}(X_i))|_{A=a}.$$

The decision associated with a decision node is a function of its parent nodes in the graph. We will have access to the value of these variables at the time of making the decision. Similarly, the utility associated with a utility node is a function of its parent nodes in the graph.

Assume that we have a finite number of discrete action choices. Then, one problem is to select the *best strategy* or function $\pi^*$ mapping each possible value of the parents of the decision node to an action choice. The best strategy is the strategy with highest expected utility. Let $X = (Z, O)$ where the variables in $O$ are parents of the decision node and $Z$ are the remaining variables. The problem of obtaining an optimal strategy reduces to obtaining, for each assignment $O = o$, the action that maximizes the value associated with the action and the assignment:

$$V_o(a) = \sum_Z P(Z, O = o \mid A = a) U(Z, O = o, A = a).$$

Note once again that computing this value requires the evaluation of a sum. For the same reasons as in the previous problem of belief inference in BNs, the exact computation of this value is intractable in general.

## 3 IMPORTANCE SAMPLING

Importance sampling provides an alternative to the exact methods for evaluating sums. Let the quantity of interest be $G = \sum_Z g(Z)$ for some real function $g$. We can turn the sum into an expectation by expressing $G = \sum_Z f(Z) (g(Z)/f(Z))$, where $f$ is a probability distribution over $Z$ satisfying, for all $Z$, $g(Z) \neq 0 \Rightarrow f(Z) \neq 0$. We call $f$ the *importance-sampling distribution*. We define the *weight function* $\omega(Z) = g(Z)/f(Z)$ which allows us to express $G = \sum_Z f(Z) \omega(Z)$. Hence, we can obtain an unbiased estimate of $G$ by obtaining $N$ samples $z^{(1)}, \ldots, z^{(N)}$ from $Z \sim f$ and computing the estimate

$$\hat{G} = \frac{1}{N} \sum_{l=1}^{N} \omega(z^{(l)}). \qquad (1)$$

We can apply this technique to the problem of belief inference in BNs. Typically, we let

$$g(Z) = P(Z, O = o)$$
$$= \prod_{i=1}^{n_1} P(Z_i \mid \text{Pa}(Z_i)) \prod_{j=1}^{n_2} P(O_j \mid \text{Pa}(O_j))\Big|_{O=o},$$
$$f(Z) = \prod_{i=1}^{n_1} P(Z_i \mid \text{Pa}(Z_i))\Big|_{O=o}, \text{ which implies}$$
$$\omega(Z) = \prod_{j=1}^{n_2} P(O_j \mid \text{Pa}(O_j))\Big|_{O=o}.$$

Note that we are defining the importance sampling distribution to be the "prior" distribution of the BN. We obtain samples from this distribution by sampling the variables in the (partial) order defined by the DAG and according to the local conditional distribution of the original BN for each variable. As we obtain samples from each variable by traversing the nodes in the graph and sampling the variable corresponding to it, if we get to a node or variable that is in the evidence set $O$, we do not sample it. Instead, we assign to it the value given by the evidence assignment $o$. Therefore, the resulting samples will be assignments to those variables that are not in the evidence set according to the "prior" distribution of the BN. We call the method resulting from this importance-sampling distribution the *traditional method*. In the context of belief inference, this method is called *likelihood-weighting (LW)* since the weight function is a "likelihood" and thus each sample is weighted by its "likelihood."

We can similarly apply this technique in the context of action selection in IDs to evaluate $V_o(a)$. In general, we let

$$g(Z) = P(Z, O = o \mid A = a) U(Z, O = o, A = a),$$
$$f(Z) = \prod_{i=1}^{n_1} P(Z_i \mid \text{Pa}(Z_i))\Big|_{O=o, A=a},$$
$$\omega(Z) = \prod_{j=1}^{n_2} P(O_j \mid \text{Pa}(O_j)) U(Z, O, A)\Big|_{O=o, A=a}.$$

In particular, for our example,

$$g(Z) = P(X_1) P(X_2 \mid X_1) P(X_3 \mid X_1) \times$$
$$P(X_6 \mid X_2, A = a) P(X_7 \mid X_3, X_6) \times$$
$$P(X_4 = x_4 \mid X_2) P(X_5 = x_5 \mid X_2, X_3) \times$$
$$U(X_7, A = a),$$
$$f(Z) = P(X_1) P(X_2 \mid X_1) P(X_3 \mid X_1) \times$$
$$P(X_6 \mid X_2, A = a) P(X_7 \mid X_3, X_6),$$
$$\omega(Z) = P(X_4 = x_4 \mid X_2) P(X_5 = x_5 \mid X_2, X_3) \times$$
$$U(X_7, A = a).$$

An important property of the estimator $\hat{G}$ is the variance of the weights associated with the importance-sampling distribution. This is

$$\text{Var}[\omega(Z)] = \sum_Z f(Z) \omega(Z)^2 - G^2.$$

Recall that $G = \sum_Z g(Z)$ by definition and assume that $g$ is a positive function. From this we can derive that the optimal or minimum-variance importance-sampling distribution is proportional to $g(Z)$:

$$f^*(Z) = g(Z) / \sum_Z g(Z). \qquad (2)$$

The weights will have zero variance in that case, since the weight function will always output our value of interest $G$. We also note that we need to avoid letting $f(Z)$ be too small with respect to $g(Z)$, since this will increase the variance. As a matter of fact, $\text{Var}[\omega(Z)] \to \infty$ as $f(Z) \to 0$ for at least one value of $Z$. This implies that we should use importance-sampling distributions with sufficiently "fat tails."

## 4 ADAPTIVE IMPORTANCE SAMPLING

The traditional method presented above uses as the importance-sampling distribution the "prior" distribution of the BN which can be far from optimal in the sense that it can have higher variance than necessary. In the case of evaluating actions in IDs, it also completely ignores potentially useful information about the utility values. Therefore, we try to learn the optimal importance-sampling distribution by adapting the current sampling distribution as we obtain samples from it.

We view the adaptive process as one of learning a distribution over the variables the sum is over to use specifically as an importance-sampling distribution. In particular, we can view this process as learning BNs from the samples just for sampling. From the expression of the optimal importance-sampling distribution given in equation 2 (and, in particular, from the factorization of the function $g$ for the different estimation problems), we can deduce that in order to be able to represent this distribution graphically using a BN we need to add arcs that connect every pair of nodes that are parents of observations and/or utility nodes, if they are not already connected. However, doing so can increase the size of the model, particularly in cases where the local conditional probabilities and the utilities have a smaller, more compact parametric representation (i.e., noise-or's). In this paper, we do not deal with this issue and instead concentrate on the problem of learning a BN with the same structure as the original BN (or ID). Hence, we only need to update the local conditional probability distributions as we obtain samples.

We can parameterize the importance-sampling distribution using a set of parameters $\Theta$. Let the indicator function $I(Z_i = k, \mathrm{Pa}(Z_i) = j \mid Z) = 1$ if the condition $Z_i = k$ and $\mathrm{Pa}(Z_i) = j$ agrees with the value assigned to $Z$; 0 otherwise. Then, we can express the importance-sampling distribution as

$$f(Z \mid \Theta) = \prod_{i=1}^{n} \prod_{j \in \Omega_{\mathrm{Pa}(Z_i)}} \prod_{k \in \Omega_{Z_i}} \theta_{ijk}^{I(Z_i=k, \mathrm{Pa}(Z_i)=j \mid Z)}, \quad (3)$$

where for each $i, j, k$, $\theta_{ijk} = P(Z_i = k \mid \mathrm{Pa}(Z_i) = j, \Theta)$. Hence, for all $i, j$, $\sum_k \theta_{ijk} = 1$, and for all $k$, $\theta_{ijk} > 0$. Note that this representation uses the assumptions of *global* and *local parameter independence* typically used in BNs. The weight function is also parameterized and defined as $\omega(Z \mid \Theta) = g(Z)/f(Z \mid \Theta)$.

### 4.1 LEARNING CRITERIA AND UPDATE RULES

In the following subsections we present different methods for updating the sampling distribution. The update rules are all based on gradient-descent. Hence, at each time $t$, we update the parameters as follows:

$$\theta^{(t+1)} \leftarrow \theta^{(t)} - \alpha(t) \nabla^p e(\theta^{(t)}). \quad (4)$$

In the update rule above, $\alpha(t)$ denotes the learning rate or the step size rule and $\nabla^p e(\Theta)$ denotes the gradient of error function $e$, appropriately projected to satisfy the constraints on $\Theta$. The methods differ in how they define $\nabla^p e(\theta^{(t)})$.

In the discussion below we denote the $N(t)$ i.i.d. samples as $z^{(t,1)}, \ldots, z^{(t,N(t))}$ drawn according to $Z \sim f(Z \mid \theta^{(t)})$. If we gather samples to estimate $G$ using many different sampling distributions, how can we combine them to get an unbiased estimate? It is sufficient to weight them using any weighting function that is independent of the sub-estimates obtained by using just the samples for one sampling distribution. For instance, the estimator

$$\hat{G}^{(T)} = \sum_{t=1}^{T} W(t) \hat{G}(\theta^{(t)}), \quad (5)$$

where $\sum_{t=1}^{T} W(t) = 1$ and $W(t) \geq 0$, for all $t$, and

$$\hat{G}(\theta^{(t)}) = \frac{1}{N(t)} \sum_{l=1}^{N(t)} \omega(z^{(t,l)} \mid \theta^{(t)}), \quad (6)$$

is unbiased as long as $W(t)$ and $\hat{G}(\theta^{(t)})$ are independent for each $t$. Letting $W(t) = 1/T$ will produce an unbiased estimate. This is the weight we use in the experiments. In general, we would like to give more weight to importance-sampling distributions with smaller variances. Assuming that the variance decreases with $t$, we would like $W(t)$ to be an increasing sequence of $t$. Note that using $W(t) \propto 1/\hat{\sigma}_t^2$, where $\hat{\sigma}_t^2$ is the sample variance at time $t$, though appealing, does not necessarily lead to an unbiased estimator since $W(t)$ and $\hat{G}(\theta^{(t)})$ are not independent.

We will consider three general strategies: minimizing variance directly, minimizing distance to *global approximations* of the optimal sampling distribution, and minimizing distance to the empirical distribution of the optimal sampling distribution based on *local approximations*. For the first two strategies, we will find that we can express the partial derivatives that form the gradient as, for all $i, j, k$,

$$\frac{\partial e(\Theta)}{\partial \theta_{ijk}} = \sum_{Z} f(Z \mid \Theta) \left[ \frac{-I(Z_i=k, \mathrm{Pa}(Z_i)=j \mid Z)}{\theta_{ijk}} \times \varphi(Z, \Theta) \right],$$

where $\varphi(Z, \Theta)$ is a function that depends on the error functions. Note that this is an expectation. Then, the methods update the parameters by estimating the value of the partial derivatives evaluated at the current setting of the parameters $\theta^{(t)}$ as

$$\frac{\partial \hat{e}(\theta^{(t)})}{\partial \theta_{ijk}} = \frac{1}{N(t)} \sum_{l=1}^{N(t)} \left[ \frac{-I(Z_i=k, \mathrm{Pa}(Z_i)=j \mid Z=z^{(t,l)})}{\theta_{ijk}^{(t)}} \times \varphi(z^{(t,l)}, \theta^{(t)}) \right].$$

#### 4.1.1 Minimizing Variance Directly

As we noted above, the optimal importance-sampling distribution for estimating $G$ is that which minimizes the variance of $\omega$. Using that as our objective, we derive a stochastic-gradient update rule for the parameters of the importance-sampling distribution. Let the error function be

$$e_{\text{Var}}(\Theta) = \text{Var}(\omega(Z \mid \Theta))$$
$$= \sum_Z f(Z \mid \Theta)\omega(Z \mid \Theta)^2 - G^2$$

The corresponding function for the gradient is

$$\varphi_{\text{Var}}(Z, \Theta) = \omega(Z \mid \Theta)^2. \quad (7)$$

Note that using this definition of $\varphi$ yields an unbiased estimate of the gradient. This is because the gradient is the expectation of a particular function and, in this case, we can always evaluate the function exactly. Hence, we can obtain an unbiased estimate by sampling from $f(Z \mid \Theta)$.

### 4.1.2 Minimizing Variance Indirectly via Approximate Global Minimization

Recall the optimal importance-sampling distribution $f^*$ for estimating $G$ given in equation 2. The update rules of the following subsection are all motivated by the idea of reducing some notion of *distance* between the current sampling distribution and this optimal sampling distribution. Note that we cannot really compute the values of the optimal distribution since that requires knowing the normalizing constant $\sum_Z g(Z) = G$ which is exactly the value we want to estimate. We approximate the optimal distribution using the current estimate of $G$ as follows

$$\hat{f}^t(Z) = g(Z)/\hat{G}^{(t)}. \quad (8)$$

In the following, we will consider four error functions, one based on the sum-squared-error and three based on versions of the *Kullback-Leibler divergence*.

If we use the $L_2$ norm or sum-squared-error function as a notion of distance between the distributions, then the error function is

$$e_{L_2}(\Theta) = \tfrac{1}{2} \sum_Z \left( f(Z \mid \Theta) - f^*(Z) \right)^2.$$

The corresponding function for the gradient is

$$\varphi_{L_2}(Z, \Theta) = f^*(Z) - f(Z \mid \Theta)$$
$$\approx f(z^{(t,l)} \mid \theta^{(t)}) \times$$
$$\left( \omega(z^{(t,l)} \mid \theta^{(t)})/\hat{G}^{(t)} - 1 \right), \quad (9)$$

where the approximation results from using $\hat{f}^t(Z)$ as defined in equation 8 as an approximation to $f^*(Z)$.

An alternative, commonly-used notion of *distance* between two probability distributions is given by the *Kullback-Leibler (KL) divergence*. This measure is not symmetric. One version of the KL divergence in this context is given by the error function

$$e_{\text{KL}_1}(\Theta) = \sum_Z f^*(Z) \log \left( f^*(Z)/f(Z \mid \Theta) \right).$$

The corresponding function for the gradient is

$$\varphi_{\text{KL}_1}(Z, \Theta) = f^*(Z)/f(Z \mid \Theta)$$
$$\approx \omega(z^{(t,l)} \mid \theta^{(t)})/\hat{G}^{(t)}. \quad (10)$$

Another version of the KL divergence is given by the error function

$$e_{\text{KL}_2}(\Theta) = \sum_Z f(Z \mid \Theta) \log \left( f(Z \mid \Theta)/f^*(Z) \right).$$

The corresponding function for the gradient is

$$\varphi_{\text{KL}_2}(Z, \Theta) = \log \left( f^*(Z)/f(Z \mid \Theta) \right) - 1$$
$$\approx \log \left( \omega(z^{(t,l)} \mid \theta^{(t)})/\hat{G}^{(t)} \right) - 1. \quad (11)$$

A "symmetrized" version of KL sometimes used is given by the error function

$$e_{\text{KL}_s}(\Theta) = \tfrac{1}{2} e_{\text{KL}_1}(\Theta) + \tfrac{1}{2} e_{\text{KL}_2}(\Theta).$$

We can obtain the partial derivatives for this error function and their approximation accordingly.

### 4.1.3 Heuristic Local Minimization Based on Empirical Distribution

The update methods in this subsection are motivated by the idea of minimizing different notions of distance between the current sampling distribution and an empirical distribution of the optimal importance-sampling distribution that we build from the samples. The hope is that the empirical distribution is a good approximation of the optimal sampling distribution. We define the empirical distribution, parameterized by $\hat{\Theta}$ locally as follows: for all $i, j, k$,

$$\hat{\theta}_{ijk}^{(t)} = \frac{\sum_{l=1}^{N(t)} I(Z_i=k, \text{Pa}(Z_i)=j \mid Z = z^{(t,l)}) \omega(z^{(t,l)} \mid \theta^{(t)})}{\sum_{l=1}^{N(t)} I(\text{Pa}(Z_i)=j \mid Z = z^{(t,l)}) \omega(z^{(t,l)} \mid \theta^{(t)})}, \quad (12)$$

if $\sum_{l=1}^{N(t)} I(\text{Pa}(Z_i) = j \mid Z = z^{(t,l)}) \omega(z^{(t,l)} \mid \theta^{(t)}) \neq 0$; $\hat{\theta}_{ijk}^{(t)} = \theta_{ijk}^{(t)}$ otherwise. We are essentially defining the empirical distribution using the samples if there are samples that can be used to define it; otherwise, we revert to the current distribution. We try to minimize the distance between the current sampling distribution and the empirical distribution locally.

Similar to the case of the previous strategies, we will find that we can express the partial derivatives that form the gradient of the error functions discussed in this subsection as, for all $i, j, k$,

$$\tfrac{\partial e'(\Theta)}{\partial \theta_{ijk}} = -\varphi'(\hat{\theta}_{ijk}, \theta_{ijk}),$$

where $\varphi'(\hat{\theta}_{ijk}, \theta_{ijk})$ is a function that depends on the error functions. Then, the methods update the parameters by estimating the value of the partial derivatives evaluated at the current setting of the parameters $\theta^{(t)}$ as

$$\tfrac{\partial \tilde{e}'(\theta^{(t)})}{\partial \theta_{ijk}} = -\varphi'(\hat{\theta}_{ijk}^{(t)}, \theta_{ijk}^{(t)}).$$

We define the *local $L_2$-norm* error function as

$$e'_{L_2}(\Theta) = \tfrac{1}{2} \sum_{i,j,k} \left(\hat{\theta}_{ijk} - \theta_{ijk}\right)^2,$$

the error function for one version of KL as

$$e'_{KL_1}(\Theta) = \sum_{i,j,k} \hat{\theta}_{ijk} \log\left(\hat{\theta}_{ijk}/\theta_{ijk}\right),$$

and the other as

$$e'_{KL_2}(\Theta) = \sum_{i,j,k} \theta_{ijk} \log\left(\theta_{ijk}/\hat{\theta}_{ijk}\right).$$

From this we obtain the corresponding functions for the gradient:

$$\begin{aligned}
\varphi'_{L_2}(\hat{\theta}_{ijk}, \theta_{ijk}) &= \hat{\theta}_{ijk} - \theta_{ijk}, \\
\varphi'_{KL_1}(\hat{\theta}_{ijk}, \theta_{ijk}) &= \hat{\theta}_{ijk}/\theta_{ijk}, \\
\varphi'_{KL_2}(\hat{\theta}_{ijk}, \theta_{ijk}) &= \log\left(\hat{\theta}_{ijk}/\theta_{ijk}\right) - 1.
\end{aligned}$$

We can obtain an update rule based on the "symmetrized" version of KL accordingly.

### 4.2 DISCUSSION OF UPDATE RULES

First, note that of all the update rules, only the one derived for $e_{Var}$ clearly uses an unbiased estimate of the gradient. It is not immediately apparent whether the update rules based on $e_{L_2}$, $e_{KL_1}$ and $e_{KL_2}$ use unbiased estimates.

Note also that the magnitude of the components of the resulting gradients are different, as suggested by their respective $\varphi$ functions. The function $\varphi_{Var}$ has magnitude proportional to the squares of the weights. The magnitudes of $\varphi_{L_2}$ and $\varphi_{KL_1}$ are linear in the weights. However, the magnitude of $\varphi_{L_2}$ is potentially smaller since it has the probability of the sample as a factor. The magnitude of $\varphi_{KL_2}$ is logarithmic in the weights.

Because we assume that $g$ is positive, the weights are positive. Hence, $\varphi_{Var}$ and $\varphi_{KL_1}$ are always positive. The function $\varphi_{L_2}$ is positive if $\omega(Z \mid \Theta)/G > 1$. Similarly, the function $\varphi_{KL_2}$ is positive if $\log(\omega(Z \mid \Theta)/G) > 1$. If $\omega(Z \mid \Theta) > G$ then the sampling distribution underestimates the value of $g$ while if $\omega(Z \mid \theta) < G$ then it overestimates the value. Therefore, the sign of $\varphi_{L_2}$ and $\varphi_{KL_2}$ depends on whether we under- or over-estimated the value of $g$. Similarly, the magnitudes of $\varphi_{Var}$, $\varphi_{L_2}$, $\varphi_{KL_1}$, and $\varphi_{KL_2}$ are related to the amount of under- or overestimation. For $\varphi_{Var}$, $\varphi_{L_2}$ and $\varphi_{KL_1}$ the magnitude is larger when the sampling distribution underestimates than when it overestimates. For $\varphi_{KL_2}$, the logarithm brings the amount of over- and underestimation to the same scale. Note that for the approximations of $\varphi_{L_2}$, $\varphi_{KL_1}$, and $\varphi_{KL_2}$, $\hat{G}$ cannot be zero, and in addition for $\varphi_{KL_2}$, $\omega(Z \mid \theta)$ cannot be zero. These conditions hold from the assumption that $g$ is positive. Note that unless we constrain the importance-sampling distribution, all the functions $\varphi_{Var}$, $\varphi_{L_2}$, $\varphi_{KL_1}$ and $\varphi_{KL_2}$ will be unbounded even if $g$ is bounded.

The local $L_2$ error function, $e'_{L_2}$, leads to an update rule for which the step size has a very intuitive interpretation as a weighting between the current importance-sampling distribution and the empirical distribution. In the case of $e'_{KL_1}$, the update direction is proportional to the ratio of the empirical distribution with respect to the current importance-sampling distribution. On the other hand, for $e'_{KL_2}$, the update direction is proportional to the logarithm of the same ratio. Note $\varphi_{KL_2}$ is not defined if at least one $\hat{\theta}^{(t)}_{ijk} = 0$. We can fix this by letting, for each $i, j, k$,

$$\hat{\theta}^{(t)}_{ijk} = \frac{\left(\sum_{l=1}^{N(t)} I(Z_i=k, Pa(Z_i)=j \mid Z=z^{(t,l)})\omega(z^{(t,l)} \mid \theta^{(t)})\right) + \theta^{(t)}_{ijk}}{\left(\sum_{l=1}^{N(t)} I(Pa(Z_i)=j \mid Z=z^{(t,l)})\omega(z^{(t,l)} \mid \theta^{(t)})\right) + 1}.$$

This is essentially imposing a Dirichlet prior with parameters equal to the current probability values on the empirical distribution parameters.

We can interpret the update rules based on local KL-divergence as adding weights to the elements of the domain of the importance-sampling distribution and renormalizing. For the version of KL-divergence with respect to the empirical distribution, we are always adding weights. We add values relative to the amount we underestimated or overestimated the magnitude of the distribution for a particular state. If we underestimated, we add weights larger than one. If we overestimated, we add weights smaller than one. For the other version of KL-divergence, due to the logarithm function, we add weight if we underestimated while we subtract weight if we overestimated. Therefore, the logarithm brings the amount of underestimation and overestimation to the same scale and adds or subtracts weight accordingly.

Note that when approximating the gradients for $e_{Var}$, $e_{L_2}$, $e_{KL_1}$ and $e_{KL_2}$, we can use as little as one sample to obtain an estimate of the gradient (i.e., $N(t) = 1$). This is not advisable for the method based on the local heuristic since the empirical distribution of the optimal sampling distribution will be highly inaccurate. Hence, the update rules based on the empirical distribution will work better when we take a larger number of samples between updates. Finally, note that when $t = 1$ and $N(t) = 1$, $\varphi_{L_2} = 0$, and therefore, the parameters will not change in the first iteration.

## 5　RELATED WORK

Different variations of importance sampling have been used for the problems discussed in this paper (See Lin and Druzdzel [1999] and the references therein). Our methods belong to the class of *forward samplers* since they sample from a distribution based on the original structure of the BN. Of these, *self-importance sampling* [Shachter and Peot, 1989, Shwe and Cooper, 1991] is the method closest to the methods proposed in this paper since it also updates the sampling distribution as it obtains information from the samples. This method has an update rule that is very similar to the one derived for $e'_{L_2}$. It updates the distribution after obtaining the empirical distribution, but the update is a weighting between the empirical distribution and the first sampling distribution used [Shwe and Cooper, 1991]. The update rule is

$$\theta_{ijk}^{(t+1)} \leftarrow (1-\alpha(t))\hat{\theta}_{ijk}^{(t)} + \alpha(t)\theta_{ijk}^{(0)}$$
$$= \theta_{ijk}^{(t)} - \alpha(t)\left(\frac{\theta_{ijk}^{(t)}}{\alpha(t)} - (1-\alpha(t))\frac{\hat{\theta}_{ijk}^{(t)}}{\alpha(t)} - \theta_{ijk}^{(0)}\right).$$

In our framework, we can think of this update rule as resulting from the error function

$$e'_{SIS}(\Theta, t) = \frac{1}{2\alpha(t)} \sum_{ijk} \left(\theta_{ijk} - \left((1-\alpha(t))\hat{\theta}_{ijk} + \alpha(t)\theta_{ijk}^{(0)}\right)\right)^2.$$

*Annealed importance sampling* [Neal, 1998] is a related technique in that it tries to obtain samples from the optimal sampling distribution. As we understand it, the user sets up a sequence of distributions, the last distribution being the optimal distribution, typically defined by Markov chains. We move from one distribution to another as we "anneal" and the sequence converges to the optimal sampling distribution. The hope is that we can get an independent sample from that distribution, then we restart the process to try to obtain another independent sample, and so on. Finally, it uses those independent samples to obtain an estimate. Notice that each "traversal" of the sequence of distributions (or Markov chains) produces a single sample. The technique is very general and we are unaware of whether it has been applied to the problems considered in this paper. We are currently investigating possible connections between our methods and this technique.

## 6 EMPIRICAL RESULTS

We implemented all of the adaptive importance-sampling methods described above. We let the learning rate $\alpha(t) = \beta/t$, where $\beta$ is a value that depends on the updating method. We need different values of $\beta$ for the different methods because of the differences in magnitude of their gradients. We impose an additional constraint on the parameters which we call the $\epsilon$-*boundary*. We require that for all $i, j, k$, $\theta_{ijk} \geq \epsilon(|\Omega_{X_i}|) = \gamma/|\Omega_{X_i}|$, where $\gamma$ is a constant factor. In our experiments, we let $\gamma = 0.1$. We do this so that our sampling distribution has "fat tails", avoiding extrema in probability and hence the possibility of infinite variance. We initialize the parameters $\theta^{(0)}$ such that the starting importance-sampling distribution is the "prior" probability distribution of the original BN. However, if one of the local conditional probability values does not satisfy the $\epsilon$-boundary constraint, we change the distribution so that it does.

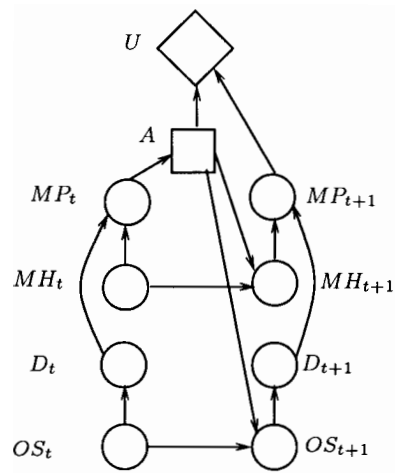

Figure 2: Graphical representation of the ID for the computer mouse problem.

In order to satisfy the constraint that for all $i, j$, $\sum_k \theta_{ijk} = 1$, we project the approximation of the gradients onto the simplex of the local conditional probability distribution. We do so by letting, for all $i, j, k$,

$$\frac{\partial^p \hat{e}(\theta)}{\partial \theta_{ijk}} \leftarrow \frac{\partial \hat{e}(\theta)}{\partial \theta_{ijk}} - \frac{1}{|\Omega_{X_i}|} \sum_{k=1}^{|\Omega_{X_i}|} \frac{\partial \hat{e}(\theta)}{\partial \theta_{ijk}}. \quad (13)$$

Note that this is not enough to guarantee that after taking a step in the projected direction, the parameters will remain in the constraint space. If, when updating a local conditional probability distribution, its respective parameters do not satisfy the constraint, we find the minimum step $\alpha'$ that will allow them to remain inside the constraint space and take a step of size $\alpha'/2$ along the gradient direction (i.e., half the distance between the current position of the parameter we are updating in the simplex and the closest point on the $\epsilon$-boundary along the gradient direction).

We tested the methods on the *computer mouse problem* [Ortiz and Kaelbling, 2000], a simple made-up ID shown in Figure 2. We added one to all the utility values presented in Ortiz and Kaelbling [2000] to make $g$ positive. We will consider the problem of obtaining the value $V_{MP_t}(A)$ for the action $A = 2$ and the observation $MP_t = 1$.

We evaluated each method by computing the *mean-squared-error (MSE)* between the true value of the expectation of interest ($V_{MP_t}(A)$) and the estimate generated using the adaptive sampling method. The first results show how the methods achieve better MSEs with fewer samples for this problem. We only show results for those methods that were the most competitive. We denote by "Var" the method based on the minimization of the variance, and by "L2", "KL1", and "KLS" the methods based on the global minimization of $L_2$, $KL_1$ and $KL_s$ respectively. For the update methods we use $N(t) = 1$ for all $t$. We take into account that the update methods have to traverse the graph once every iteration to update the parameters relevant to the sample taken. To compensate for this time, we allow the estimate based on LW to use twice as many samples. Figure 3 shows the results. The graph shows the average MSE over 40 runs as a function of the total number of samples taken (times 2 for LW) by the methods. We note that Var and L2 achieve better MSEs than LW and converge to them faster. With significance level 0.005 we can state (individually) for each total number of samples $N = 50, 150, 250$, that Var and L2 (individually) are better with respect to MSE than LW. Also, for $N = 250$, KLS is better than LW.

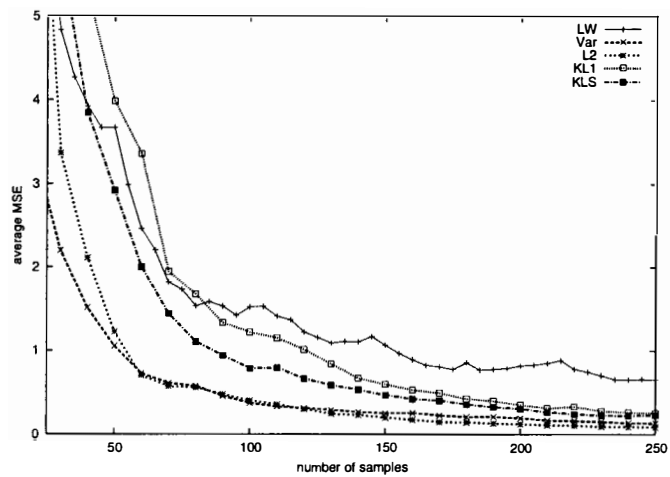

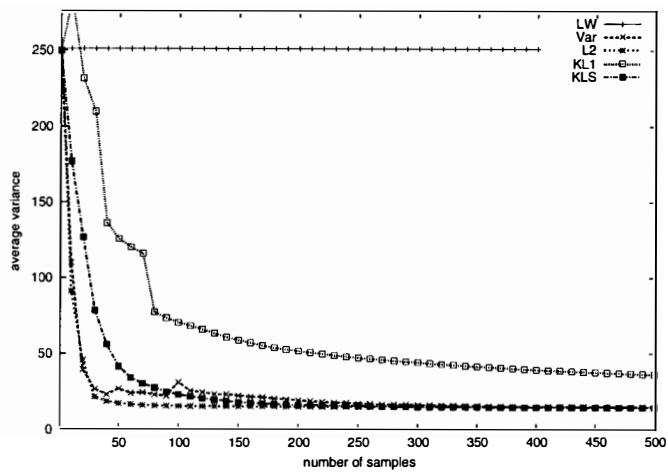

Figure 3: Average mean squared error, over 40 runs, as a function of the number of samples taken. We allow LW twice as many samples.

Figure 4: Average of the true variance of the weight function, over 40 runs, as a function of the total number of samples taken.

We also ran the methods with $N(t) = 50$, including the local heuristic methods. They were only competitive after a larger total number of samples ($N > 150$). Although further analysis is necessary, we would like to convey some general observations. We believe that in general there is a tradeoff in the setting of $N(t)$ and $\beta$. We note that, of the updates based on the two KL versions, KL1 typically performs better than KL2. We believe this is because the error function $e_{KL_1}$ is defined with respect to the optimal sampling distribution while $e_{KL_2}$ is with respect to the current sampling distribution. KLS seems to perform better than both. L2 is more stable than any of the other methods, suggesting further theoretical analysis which we are currently undertaking. Several possible reasons for this behavior are (1) the variance of the gradient might be smaller than in other cases, (2) the error function is bounded, and/or (3) the error surface might be smoother than in other cases. We conjecture that L2 converges to a stationary point of $e_{L_2}$.

The second result shows that the update methods indeed lead to importance-sampling distributions with smaller variance relatively quickly for this problem. Figure 4 shows a graph of the true variance of the sampling distribution learned using the different update methods as a function of the total number of samples used. The horizontal line shows the variance associated with the sampling distribution used by LW (i.e., the "prior" distribution of the original BN).

These experiments are all carried out on a single problem. Although they must clearly be extended to a variety of larger problems, they indicate that adaptive importance-sampling methods, particularly those that minimize variance and the $L_2$ norm, can lead to significant improvements in the efficiency of sampling as a method for computing large expectations.

## Acknowledgments

The dynamic weighting scheme and the $1/\sigma^2$ recommendation in Section 4.1 and the $\epsilon$-boundary in Section 6 were independently developed by Jian Cheng and Marek Druzdzel. Both heuristics are reported in a manuscript that the first author saw while he was working on this paper.

We would like to thank Milos Hauskrecht, Thomas Hofmann, Kee-Eung Kim and Thomas Dean for many discussions and feedback. Also, our implementation uses some of the functionality of the *Bayes Net Toolbox for Matlab* [Murphy, 1999], for which we thank Kevin Murphy. We would also like to thank the anonymous reviewers for their insightful comments.

Luis E. Ortiz was supported in part by an NSF Graduate Fellowship and in part by NSF IGERT award SBR 9870676. Leslie Pack Kaelbling was supported in part by a grant from NTT and in part by DARPA Contract #DABT 63-99-1-0012.

# References

John M. Charnes and Prakash P. Shenoy. A forward Monte Carlo method for solving influence diagrams using local computation. School of Business, University of Kansas, Working Paper No. 273, August 1999.

Gregory F. Cooper. The computational complexity of probabilistic inference using Bayesian belief networks. *Artificial Intelligence*, 42:393–405, 1990.

Robert Fung and Kuo-Chu Chang. Weighting and integrating evidence for stochastic simulation in Bayesian networks. In *Proceedings of the Fifth Workshop on Uncertainty in Artificial Intelligence*, pages 112–117, 1989.

John Geweke. Bayesian inference in econometric models using Monte Carlo integration. *Econometrica*, 57(6): 1317–1339, November 1989.

Yan Lin and Marek Druzdzel. Stochastic sampling and search in belief updating algorithms for very large Bayesian networks. In *Working Notes of the AAAI Spring Symposium on Search Techniques for Problem Solving Under Uncertainty and Incomplete Information*, pages 77–82, Stanford, California, March 1999. Stanford University. Available from http://www.pitt.edu/~druzdzel/publ.html.

Kevin P. Murphy. Bayes net toolbox for Matlab, 1999. Available from http://www.cs.berkeley.edu/~murphyk/Bayes/bnt.html.

Radford M. Neal. Annealed importance sampling. Technical Report 9805, Department of Statistics, University of Toronto, Toronto, Ontario, Canada, September 1998. Available from http://www.cs.utoronto.ca/~radford/.

Luis E. Ortiz and Leslie Pack Kaelbling. Sampling methods for action selection in influence diagrams. In *Proceedings of the Seventeenth National Conference on Artificial Intelligence*, 2000. Forthcomming.

Ross D. Shachter and Mark A. Peot. Simulation approaches to general probabilistic inference on belief networks. In *Proceedings of the Fifth Workshop on Uncertainty in Artificial Intelligence*, pages 311–318, 1989.

Michael Shwe and Gregory Cooper. An empirical analysis of likelihood-weighting simulation on a large, multiply connected medical belief network. *Computers and Biomedical Research*, 24:453–475, 1991.